\documentclass[10pt,twocolumn,letterpaper]{article}

\usepackage[pagenumbers]{cvpr} 
\usepackage{booktabs}
\usepackage{multirow}
\usepackage{adjustbox}

\usepackage{array}
\usepackage{colortbl}
\usepackage{multirow}
\usepackage{siunitx}

\usepackage{float}

\definecolor{cvprblue}{rgb}{0.21,0.49,0.74}
\usepackage[pagebackref,breaklinks,colorlinks,allcolors=cvprblue]{hyperref}

\def\model{\textit{CAMS}}
\def\graylevel{gray!20}

\title{CAMS: Towards Compositional Zero-Shot Learning via Gated Cross-Attention and Multi-Space Disentanglement}

\author{Pan Yang$^{1}$\thanks{Co-first author.}, Cheng Deng$^{2}$\footnotemark[1], Jing Yang$^{1}$\thanks{Corresponding author.}, Han Zhao$^{2}$\footnotemark[2],
Yun Liu$^{3}$, Yuling Chen$^{1}$, \\ Xiaoli Ruan$^{1}$, Yanping Chen$^{1}$\\
{$^1$The State Key Laboratory of Public Big Data, Guizhou University}\\
{$^2$Shanghai Jiao Tong University} \ \ {$^3$Nankai University} 
\\{\tt\small gs.pyang24@gzu.edu.cn; jyang23@gzu.edu.cn\footnotemark[2]}
}
\begin{document}
\maketitle
\begin{abstract}
Compositional zero‑shot learning (CZSL) aims to learn the concepts of attributes and objects in seen compositions and to recognize their unseen compositions. Most Contrastive Language-Image Pre-training (CLIP)‑based CZSL methods focus on disentangling attributes and objects by leveraging the global semantic representation obtained from the image encoder. However, this representation has limited representational capacity and do not allow for complete disentanglement of the two. To this end, we propose CAMS, which aims to extract semantic features from visual features and perform semantic disentanglement in multidimensional spaces, thereby improving generalization over unseen attribute–object compositions. 
Specifically, CAMS designs a Gated Cross-Attention that captures fine-grained semantic features from the high-level image encoding blocks of CLIP through a set of latent units, while adaptively suppressing background and other irrelevant information. Subsequently, it conducts Multi-Space Disentanglement to achieve disentanglement of attribute and object semantics. Experiments on three popular benchmarks (MIT‑States, UT‑Zappos, and C‑GQA) demonstrate that CAMS achieves state‑of‑the‑art performance in both closed‑world and open‑world settings. The
code is available at \href{https://github.com/ybyangjing/CAMS}{https://github.com/ybyangjing/CAMS}.
\end{abstract}    
\begin{figure}[t]
  \centering
  \includegraphics[width=1\linewidth]{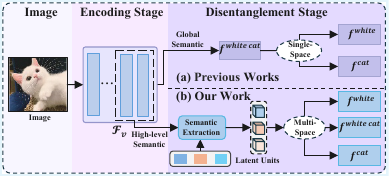}  
  \caption{Comparison between \model \ and previous approaches. (a) Existing methods perform attribute–object disentanglement in a single-space using the global semantic representation. (b) \model \ introduces latent units to extract more fine-grained semantic features from high-level encoding blocks, and then employs these features to model separate representation in multi-space, thereby achieving more effective attribute–object disentanglement.}
\label{fig:1}
\end{figure}
\section{Introduction}
\label{sec:intro}
Even if humans have never seen a ``black flower'' before, they can quickly recognize this composition by relying on their compositional generalization ability---the capacity to decompose the attribute (``black'') and object (``flower''), and compose them for inference. Compositional zero-shot learning (CZSL)~\cite{c2} draws inspiration from this ability, leveraging knowledge of seen attribute--object compositions to infer their unseen compositions.

Current dominant CZSL solutions have evolved from aligning visual and textual representations using pre-trained image encoders and word embedding models~\cite{c8,c6,c9,c4} to a new paradigm that employs large-scale pre-trained vision-language models (e.g., CLIP~\cite{c10}) as the backbone architecture~\cite{c11,c14,c18,c19}. This latter approach operates by inserting learnable attribute and object embeddings into a prompt template (e.g., ``a photo of [attribute] [object]''). This prompt is then projected into the text space using the pre-trained text encoder. The resulting textual representation is aligned with the image representation obtained from the pre-trained image encoder in a cross-modal fashion. Owing to the powerful prior knowledge embedded in these models, this method has achieved impressive results on CZSL tasks.

However, \ CLIP-based framework also introduces significant new challenges.
\textbf{(1) Lack of fine-grained semantic awareness.} CLIP-based methods emphasize overall semantic consistency but overlook subtle differences in fine-grained semantics within images. Such as “Patent.Leather Heels” and “Leather Heels” exhibit subtle differences in material texture. \textbf{(2) The global representation space is highly entangled.} They attempt to directly disentangle attributes and objects in the global semantic representation space, but overlook the fact that attributes and objects share the same space, which limits the model’s generalization to unseen compositions. To address the above challenges, we propose Towards Compositional Zero-Shot Learning via Gated \textbf{C}ross-\textbf{A}ttention and \textbf{M}ulti-Space \textbf{D}isentanglement (\model), a novel fine-grained semantic disentanglement framework. We first extract semantic features from visual inputs, and then model the attribute, object, and composition spaces to achieve effective disentanglement between attributes and objects. Specifically, our novel design includes the following steps. (1) \textbf{Encoding Stage}: Unlike the method in \cref{fig:1}(a), we perform disentanglement using high-level semantic features. (2) \textbf{Disentanglement Stage}: We introduce a set of latent units to extract semantic features, which provide stronger fine-grained representational capability, thereby facilitating subsequent attribute-object disentanglement---as illustrated in \cref{fig:1}(b). Our main contributions are summarized as follows:
\begin{itemize}
    \item  We propose a novel semantic disentangled framework for CZSL, which leverages high-level semantic features to address the insufficient disentanglement of attribute and object semantic concepts caused by the limited capacity of CLIP’s image encoder to represent the global semantic representation.
    \item Additionally, we design a Gated Cross-Attention, which enables the latent units to adaptively extract visual semantic features.
    \item We further apply Multi-Space Disentanglement to the semantic features and align them with prompt representations, which substantially enhances the model’s generalization ability to unseen compositions.
    \item In both closed-world and open-world settings, \model \ is systematically evaluated on three main CZSL datasets and achieves SOTA performance, with HM and AUC improving by up to \emph{+9.3\%} and \emph{+12.4\%}, respectively, over existing SOTA methods.
\end{itemize}
\section{Related work}
\subsection{Compositional Zero-Shot Learning} The goal of Compositional Zero-Shot Learning (CZSL) is to utilize known attribute-object compositions to recognize unseen ones. Early methods follow two paradigms: (1) aligning textual and visual representations of full compositions in a shared space \cite{c20,c1,c25,c26,c3}; (2) decoupling attributes and objects via separate classifiers \cite{c6,c9,c27,c28,c55}. Currently, since vision-language models perform excellently in downstream tasks, some studies have employed them as the backbone architecture for CZSL methods, representing a new paradigm. CSP \cite{c12} uses prompt engineering with trainable attribute-object embeddings. Troika \cite{c11} extends CLIP to a triple-path framework modeling attributes, objects, and compositions independently.

Despite these advances, they rely on a single global semantic representation for disentangling attributes and objects, overlooking its limited capacity and hindering fine-grained semantic understanding.

\subsection{Gating in Neural Networks} Gating mechanisms are widely used in deep neural networks to control information flow and improve learning dynamics. In early RNNs, LSTM \cite{c31} and GRU \cite{c32} employ gating units to handle long-term dependencies, effectively alleviating vanishing/exploding gradients. More recently, gating has been integrated into Transformers \cite{c34} to enhance representation capacity and inter-layer flow. For instance, SwiGLU \cite{c35} applies gating in feedforward layers for better nonlinearity, while RetNet \cite{c37} uses it in token-mixing modules for efficient and stable sequence modeling. Recent studies \cite{c39} also show that gating in attention mechanisms boosts performance.

We extend this concept to cross-attention by introducing a gating mechanism, enabling selective control of information flow during feature interaction. This promotes precise modeling and fine-grained feature extraction.

\subsection{Semantic Disentanglement} Semantic disentanglement separates entangled factors (e.g., color, shape) in visual tasks, enabling independent representation in subspaces to boost transferability and generalization in novel scenarios \cite{c40,c41}. Early work \cite{c42} proposed theoretical metrics for semantic overlap and used information theory or matrix decomposition for separation. With deep learning advances, decoupling is now achieved via learnable modules in end-to-end networks across tasks like Unsupervised \cite{c43,c44}, Few-Shot \cite{c45}, and CZSL. In CZSL, composition concepts are decomposed into independent attribute and object representations. Such as, CDS-CZSL \cite{c18} employs context-aware, diversity-guided specificity learning to favor attributes conveying greater semantic information. CLUSPRO \cite{c17} refines visual prototypes via self-supervised clustering.

They usually perform disentanglement in single-space, which is often constrained by limited representational capacity. In contrast, our method models fine-grained semantic features across multi-space, achieving more efficient semantic disentanglement.
\section{Fundamental Method}
\subsection{Task Formulation}
In CZSL, similar to image classification, each image $I$ is associated with a label \textit{y}. Unlike standard classification, the label \textit{y} in CZSL is composed of a pair—an attribute ($y_{attr} \in A $) and an object ($y_{obj} \in O $), $y=(y_{attr},y_{obj}) \in Y$. The goal of CZSL is to learn the relevant visual concepts of attributes and objects from seen compositions $y \in Y_{s}$, to generalize them to unseen compositions $y \in {Y_{u}}$, where $Y_{s} \cap {Y_{u}}= \oslash $. If the test set compositions $Y_{t}$ satisfies $Y_{t} \cap {Y_{s}} = \oslash $ , the setting is referred to as traditional CZSL. Conversely, if $Y_{t}$ satisfy $Y_{t} \cap {Y_{s}} \neq \oslash $ , it is referred to as Closed-World CZSL (CW-CZSL). When the test set compositions cover the full space $Y_{t} \equiv A \times O$, the setting is known as Open-World CZSL (OW-CZSL). In this case, $|Y_{t}| \gg |Y_{s}|$ requires the model to generalize from a small number of seen compositions to a much larger number of unseen compositions, making the task extremely challenging.

\subsection{Representations Encoding}
In CZSL task, it is particularly important to extract rich multimodal representations for image-text alignment. We use the pre-trained CLIP image encoder based on the Vision Transformer (ViT) \cite{c47} architecture, together with the text encoder, as the backbone networks for the image encoder $\mathcal{F}_v$ and the text encoder $\phi_{t}$, respectively. Both encoders consist of multiple Transformer blocks and are capable of learning multimodal representations. On the visual side, given an input image $x \in \mathbb{R} ^{\textit{H}\times\textit{W}  \times 3}$, the image encoder extracts the global semantic representation $\textit{f}^g=\mathcal{F}_v(x)$. On the textual side, We follow existing CZSL prompt tuning methods \cite{c11}, passing the attribute label set $A=\{a_i\}_{i=1}^{n}$ and the object label set $O=\{o_i\}_{i=1}^{m}$ through the pretrained CLIP text‐embedding encoder \cite{c10} to obtain the attribute embedding features set $E_{A}$ and the object embedding features set $E_{O}$:
\begin{equation}\label{eq1}
    E_{A} = \{e_a^{i}\}_{i=1}^{n},\quad
    E_{O} = \{e_o^{i}\}_{i=1}^{m},
\end{equation}
where $e_a$ and $e_o$ denote the attribute and object embedding features, respectively, and $n$ and $m$ are the total numbers of attributes and objects in the dataset,respectively. For every candidate attribute, object, and attribute–object pair, we construct the following soft prompts: $T_{i}^{a}=[\theta_{1},\theta_{2},...,\theta_{r},e_{a}^{i}], T_{j}^{o}=[\theta_{1},\theta_{2},...,\theta_{r},e_{o}^{j}],$ and $T_{i,j}^{c}=[\theta_{1},\theta_{2},...,\theta_{r},e_{a}^{i},e_{o}^{j}]$, where $\{\theta_{i}\}_{i=1}^{r}$ are learnable prefix token embeddings initialized from the phrase “a photo of”. Finally, each soft prompt is passed through the text encoder $\phi_{t}$ to produce its corresponding prompt representation:
\begin{equation}\label{eq2}
\begin{aligned}
t_{\xi}^{z} = \phi_{t}(T_{\xi}^{z}), \quad z \in \{a, o, c\},
\end{aligned}
\end{equation}
where $\xi$ denotes the index of the corresponding set.
\section{CAMS: A Disentanglement Framework}
\begin{figure*}[t]
\centering
\includegraphics[width=1\textwidth]{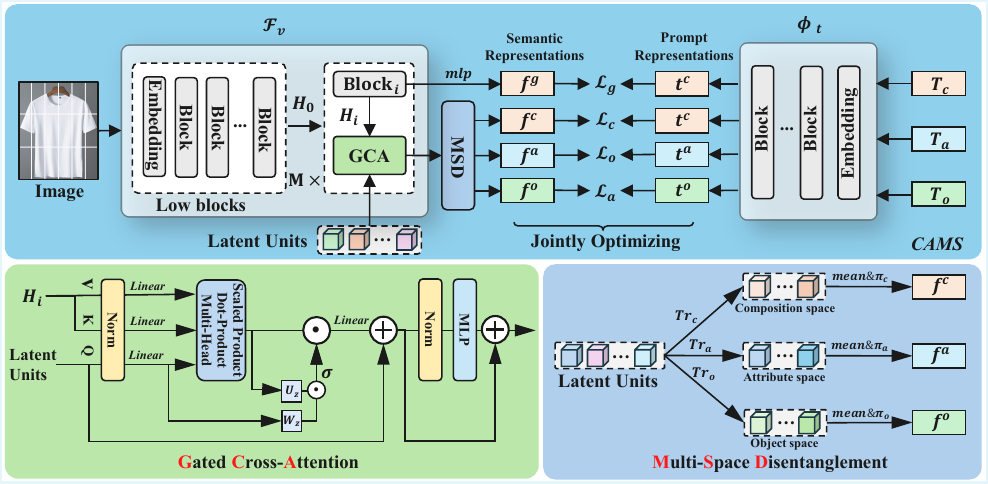} 
\caption{Illustration of \model. First, on the visual side $\mathcal{F}_v$, the global semantic representation $f^g$ of the image is is extracted through the image encoder. Simultaneously, in the last $M$ layers of the image encoder, we introduce a set of Latent Units to extract high-level semantic features of the image using GCA. The semantic features obtained by the Latent Units are then fed into MSD to produce attribute $f^a$, object $f^o$, and compositional $f^c$ semantic representations. Second, on the textual side $\phi_{t}$, we extract attribute $t^a$, object $t^o$, and composition \ $t^c$ prompt representations through a three-branch prompting mechanism using the CLIP text encoder. Finally, the semantic representations and prompt representations are optimized and aligned via cross-entropy. Where $Tr_a$, $Tr_o$, $Tr_c$ denote the Transformer encoders, and $\pi_a$, $\pi_o$, $\pi_c$ are  projection layers. $W_z$ and $U_z$ are the parameter matrices and $\sigma$ is sigmoid function.}
\label{fig:2}
\end{figure*}

Fully decoupling attributes and objects from their visual representations and learning the corresponding semantic concepts can enhance a model's generalization ability for unseen compositions. On the basis of this insight, we propose \model. \cref{fig:2} illustrates the overall architecture of \model, which comprises four main components: the image encoder $\mathcal{F}_v$, the Gated Cross-Attention, the Multi-Space Disentanglement, and the text encoder $\phi_t$. The Gated Cross-Attention is designed to capture fine-grained semantic features from images. Multi-Space Disentanglement is responsible for decoupling attributes and objects. Soft prompt corresponding to attributes, objects, and their compositions are fed into the $\phi_t$ to extract prompt representations. Finally, cross-entropy loss functions are employed to align semantic and prompt representations across each branch.

\subsection{Gated Cross-Attention}
The global semantic representation extracted by the CLIP image encoder backbone tend to emphasize holistic semantic alignment at the expense of fine‑grained detail. Given that ViT \cite{c47} intrinsically focus on visual details in low layers while capturing semantic information in high layers, to fully capture fine-grained visual semantic information, we design a gated cross-attention and draw inspiration from the Perceiver \cite{c50} by introducing a set of latent units $\textit{Q} \in \mathbb{R}^{K\times d_{v} }$ for high-level semantic features extraction. These latent units interact with the visual features of the high-level image encoder blocks, enabling efficient capture of deep latent semantic features. As shown in the left part of \cref{fig:2}, an input image $x$ is first passed through the low Transformer encoder blocks of the image encoder $\mathcal{F}_v$ , producing initial hidden-state features $\textit{H}_0 \in \mathbb{R}^{P\times d_{v}}$. We denote the output of the $\textit{i}$ th layer among the remaining $M$ Transformer encoder blocks as $\textit{H}_\textit{i} \in \mathbb{R}^{P\times d_{v}}$. 

Specifically, we first adopt a scaled dot-product attention mechanism \cite{c34}, treating $\textit{H}_{i}$ as the key and value: $\textit{Q}$ as the query:
\begin{equation}\label{eq3}
  q = QW_q,\quad K =H_iW_k,\quad V=H_iW_v,
\end{equation}
where $W_{q}, W_{k}$, and $W_{v}$ are the parameter matrices for queries, keys, and values respectively. Attention computes the relevance of queries to keys, normalizes these scores via softmax, and aggregates the values according to the resulting weights as
\begin{equation}\label{eq4}
q_1=\operatorname{Attention}(q,K,V)=\operatorname{Softmax}\left(\frac{qK^\top}{\sqrt{d_{k}}}\right)V.
\end{equation}
In practice, we introduce a multi-head attention. To effectively aggregate useful fine-grained semantic features, we introduce a gating mechanism following the attention mechanism to control the information flow:
\begin{equation}\label{eq5}
q_2 = q_1 \odot \operatorname{sigmoid}(W_zq \odot U_zq_1),
\end{equation}
where $W_{z}$ and $U_{z}$ are the parameter matrices; $\odot$ is hadamard product.
And then through an output layer:
\begin{equation}\label{eq6}
     Q' = q_2W_o,
\end{equation}
where $W_{o}$ is the parameter matrix. Subsequently, we introduce a feedforward neural network (FFN), which utilizes a multilayer perceptron (MLP) to refine the interaction feature and employs residual connections to generate the final output, thereby enhancing the features representation:
\begin{equation}\label{eq7}
 \tilde{Q}=Q^{\prime} + \operatorname{MLP}\left(\operatorname{LayerNorm}(Q^{\prime})\right).
\end{equation}
In practice, the gated cross-attention mechanism weights used across all higher layers are shared, which is similar to the structure of the RNN. We hypothesize that this gating mechanism acts as an “update gate” within the model, controlling the process of information updating and retention.
\subsection{Multi-Space Disentanglement}
In previous work on disentangling attribute and object semantics, the global semantic representations extracted by an image encoder are typically directly split into attribute and object representations. However, this split still resides within the same space and does not effectively model independent representation spaces for attributes and objects. In our method, we use latent units that have extracted fine-grained semantic features as the features to be disentangled. The disentanglement is performed in a multi-dimensional feature space, projecting these fine-grained features into three separate spaces—the attribute space, the object space, and their composition space—thereby achieving a more precise semantic decomposition. We adopt Transformer encoders \cite{c34} as our independent feature space modelers. Specifically, the latent units are first fed into their respective feature space modelers for spatial mapping:
\begin{equation}\label{eq8}
\tilde{Q}_\xi = \operatorname{Trs_{\xi}}(\tilde{Q}), \quad \xi \in \{a,o,c\},
\end{equation}
where $\operatorname{Trs_{a}}$, $\operatorname{Trs_{o}}$, and $\operatorname{Trs_{c}}$ are all Transformer encoders.
Then, global representations of the features within each space are obtained through average pooling, resulting in more stable and expressive semantic features:
\begin{equation}\label{eq9}
F^\xi = \operatorname{Mean}(\tilde{Q}_\xi), \quad \xi \in \{a,o,c\}.
\end{equation}
Finally, we obtain the corresponding representations of each branch after modeling through a projection layer:
\begin{equation}\label{eq10}
f^\xi = \pi_{\xi}(F^\xi),\quad \xi \in \{a,o,c\},
\end{equation}
where $\pi_{a}$, $\pi_{o}$, and $\pi_{c}$ denote the projection layers for the attribute, the object, and their composition, respectively.
\subsection{Training and Inference}
By combining prompt representations and semantic representations from each branch, the label probabilities for the image’s attribute branch $a$, object branch $o$, composition branch $c$, and global branch $g$ can be computed separately as
\begin{equation}\label{eq11}
\begin{aligned}
p(z_\xi \mid x) = \frac{\exp(f^z \cdot t_\xi^z / \tau)}{\sum_{k=1}^{|Z|} \exp(f^z \cdot t_k^z / \tau)},\\
z \in \{a, o, c, g\}, Z = 
\begin{cases}
A, & z = a \\
O, & z = o\\
Y_s, & z = c\\
Y_s, & z = g
\end{cases}.
\end{aligned}
\end{equation}
During the training process, the cross-entropy loss of each branch encourages the model to explicitly recognize the semantic role corresponding to it, described as
\begin{equation}\label{eq12}
\mathcal{L}_{\xi} = -\frac{1}{|\chi|} \sum_{x\in \chi} \operatorname{log}\,p(\xi|x), \quad \xi \in \{a,o,c,g\},
\end{equation}
the total loss is defined as follows:
\begin{equation}\label{eq13}
\mathcal{L} = \sum{\alpha_\xi\mathcal{L}_{\xi}},\quad \xi \in \{a,o,c,g\}.
\end{equation}
In inference, the image $x$ is passed through \model to produce the attribute prediction score, the object prediction score, the global prediction score and the composition prediction score. These four scores are then linearly combined to form the final prediction for the pair:
\begin{equation}\label{eq14}
s = \beta \cdot p(g_{i,j}|x) + (1 - \beta) \cdot p(c_{i,j}|x) + p(a_i|x) \cdot p(o_j|x),
\end{equation}
where $\beta$ is a weighting coefficient used to control the relative contribution of the global and composition branches. Top-scoring compositions are the model's prediction.
\begin{table*}[h]
\centering
\caption{Comparison results with SOTA methods in \textbf{Closed-World} setting on MIT-States, UT-Zappos, and C-GQA. ``S'', ``U'', ``HM'', and ``AUC'' stand for best Seen accuracy, best Unseen accuracy, best Harmonic Mean, and Area Under the Curve, respectively. The best results are in \textbf{bold}, and the second-best results are marked with an \underline{underline}. }
\setlength{\tabcolsep}{2mm}
\begin{tabular}{l cccc cccc cccc}
\toprule
\multirow{2}{*}{Method} & \multicolumn{4}{c}{MIT-States} & \multicolumn{4}{c}{UT-Zappos} & \multicolumn{4}{c}{C-GQA} \\
\cmidrule(lr){2-5} \cmidrule(lr){6-9} \cmidrule(lr){10-13}
& S & U & HM & AUC & S & U & HM & AUC & S & U & HM & AUC \\
\midrule
CSP \cite{c12} {\color{gray}\footnotesize [ICLR23]} & 46.6 & 49.9 & 36.3 & 19.4 & 64.2 & 66.2 & 46.6 & 33.0 & 28.8 & 26.8 & 20.5 & 6.2 \\
DFSP \cite{c15} {\color{gray}\footnotesize [CVPR23]} & 46.9 & 52.0 & 37.3 & 20.6 & 66.7 & 71.7 & 47.2 & 36.9 & 38.2 & 32.9 & 27.1 & 10.5 \\
CAILA \cite{c46} {\color{gray}\footnotesize [WACV24]} & 51.0 & 53.9 & 39.9 & 23.4 & 67.8 & 74.0 & 57.0 & 44.1 & 43.9 & 38.5 & 32.7 & 14.8 \\
CDS-CZSL \cite{c18} {\color{gray}\footnotesize [CVPR24]} & 50.3 & 52.9 & 39.2 & 22.4 & 63.9 & 74.8 & 52.7 & 39.5 & 38.3 & 34.2 & 28.1 & 11.1 \\
Troika \cite{c11} {\color{gray}\footnotesize [CVPR24]} & 49.0 & 53.0 & 39.3 & 22.1 & 66.8 & 73.8 & 54.6 & 41.7 & 41.0 & 35.7 & 29.4 & 12.4 \\
PLID \cite{C30} {\color{gray}\footnotesize [ECCV24]} & 49.7 & 52.4 & 39.0 & 22.1 & 67.3 & 68.8 & 52.4 & 38.7 & 38.8 & 33.0 & 27.9 & 11.0 \\
RAPR \cite{c14} {\color{gray}\footnotesize [AAAI24]} & 50.0 & 53.3 & 39.2 & 22.5 & 69.4 & 72.8 & 56.5 & 44.5 & \underline{45.6} & 36.0 & 32.0 & 14.4 \\
MSCI \cite{c19} {\color{gray}\footnotesize [IJCAI25]} & 50.2 & 53.4 & 39.9 & 22.8 & 67.4 & 75.5 & \textbf{59.2} & 45.8 & 42.4 & 38.2 & 31.7 & 14.2 \\
Logic-CSP \cite{c51} {\color{gray}\footnotesize [CVPR25]} & 48.5 & 51.0 & 37.4 & 20.6 & 65.1 & 69.8 & 51.0 & 38.2 & 31.9 & 28.9 & 23.1 & 7.7 \\
Logic-Troika \cite{c51} {\color{gray}\footnotesize [CVPR25]} & 50.8 & 53.9 & 40.5 & 23.4 & 69.6 & 74.9 & 57.8 & 45.8 & 44.4 & \underline{39.4} & \underline{33.3} & \underline{15.3} \\
CLUSORO \cite{c17} {\color{gray}\footnotesize [ICLR25]} & \underline{52.1} & \underline{54.0} &\underline{40.7} & \underline{23.8} & \underline{70.7} & \underline{76.0} & \underline{58.5} & \underline{46.6} & 44.3 & 37.8 & 32.8 & 14.9 \\
\rowcolor{\graylevel}
\textbf{\model\ (Ours)} & \textbf{53.0} & \textbf{54.3} & \textbf{41.0} & \textbf{24.2} & \textbf{71.6} & \textbf{76.6} & \underline{58.5} & \textbf{47.4} & \textbf{46.7} & \textbf{42.1} & \textbf{36.4} & \textbf{17.4} \\
\bottomrule
\end{tabular}
\label{tab:1}
\end{table*}
\begin{table*}[h]
\centering
\caption{Comparison results with SOTA methods in \textbf{Open-World} setting on MIT-States, UT-Zappos, and C-GQA.}
\setlength{\tabcolsep}{2mm}
\begin{tabular}{l cccc cccc cccc}
\toprule
\multirow{2}{*}{Method} &
\multicolumn{4}{c}{MIT-States} & \multicolumn{4}{c}{UT-Zappos} & \multicolumn{4}{c}{C-GQA} \\
\cmidrule(lr){2-5} \cmidrule(lr){6-9} \cmidrule(lr){10-13}
& S & U & HM & AUC & S & U & HM & AUC & S & U & HM & AUC \\
\midrule
CSP \cite{c12} {\color{gray}\footnotesize [ICLR23]} & 46.3 & 15.7 & 17.4 & 5.7 & 64.1 & 44.1 & 38.9 & 22.7 & 28.7 & 5.2 & 6.9 & 1.2 \\
DFSP \cite{c15} {\color{gray}\footnotesize [CVPR23]} & 47.5 & 18.5 & 19.3 & 6.8 & 66.8 & 60.0 & 44.0 & 30.3 & 38.3 & 7.2 & 10.4 & 2.4 \\
CAILA \cite{c46} {\color{gray}\footnotesize [WACV24]} & 51.0 & 20.2 & 21.6 & 8.2 & 67.8 & 59.7 & 49.4 & 32.8 & 43.9 & 8.0 & 11.5 & 3.1 \\
CDS-CZSL \cite{c18} {\color{gray}\footnotesize [CVPR24]} & 49.4 & 21.8 & 22.1 & 8.5 & 64.7 & 61.3 & 48.2 & 32.3 & 37.6 & 8.2 & 11.6 & 2.7 \\
Troika \cite{c11} {\color{gray}\footnotesize [CVPR24]} & 48.8 & 18.7 & 20.1 & 7.2 & 66.4 & 61.2 & 47.8 & 33.0 & 40.8 & 7.9 & 10.9 & 2.7 \\
PLID \cite{C30} {\color{gray}\footnotesize [ECCV24]} & 49.1 & 18.7 & 20.0 & 7.3 & 67.6 & 55.5 & 46.6 & 30.8 & 39.1 & 7.5 & 10.6 & 2.5 \\
RAPR \cite{c14} {\color{gray}\footnotesize [AAAI24]} & 49.9 & 20.1 & 21.8 & 8.2 & 69.4 & 59.4 & 47.9 & 33.3 & \underline{45.5} & \underline{11.2} & \underline{14.6} & \underline{4.4} \\
MSCI \cite{c19} {\color{gray}\footnotesize [IJCAI25]} & 49.2 & 20.6 & 21.2 & 7.9 & 67.4 & 63.0 & 53.2 & 37.3 & 42.0 & 10.6 & 13.7 & 3.8 \\
Logic-CSP \cite{c51} {\color{gray}\footnotesize [CVPR25]} & 48.4 & 18.6 & 19.1 & 6.9 & 65.3 & 50.8 & 42.5 & 26.3 & 31.5 & 5.2 & 7.2 & 1.4 \\
Logic-Troika \cite{c51} {\color{gray}\footnotesize [CVPR25]} & 50.7 & 21.4 & 22.4 & 8.7 & 69.6 & 63.7 & 50.8 & 36.2 & 43.7 & 9.3 & 12.6 & 3.4 \\
CLUSORO \cite{c17} {\color{gray}\footnotesize [ICLR25]} & \textbf{51.2} & \underline{22.1} & \textbf{23.0} & \textbf{9.3} & \underline{71.0} & \underline{66.2} & \underline{54.1} & \underline{39.5} & 41.6 & 8.3 & 11.6 & 3.0 \\
\rowcolor{\graylevel}
\textbf{\model\ (Ours)} & \underline{50.9} & \textbf{22.3} & \underline{22.8} & \underline{9.1} & \textbf{71.2} & \textbf{66.4} & \textbf{54.5} & \textbf{40.3} & \textbf{46.2} & \textbf{11.7} & \textbf{15.3} & \textbf{4.8} \\
\bottomrule
\end{tabular}
\label{tab:2}
\end{table*}
\begin{figure*}[htbp]
  \centering
  \begin{subfigure}[b]{0.25\textwidth}
    \includegraphics[width=\linewidth]{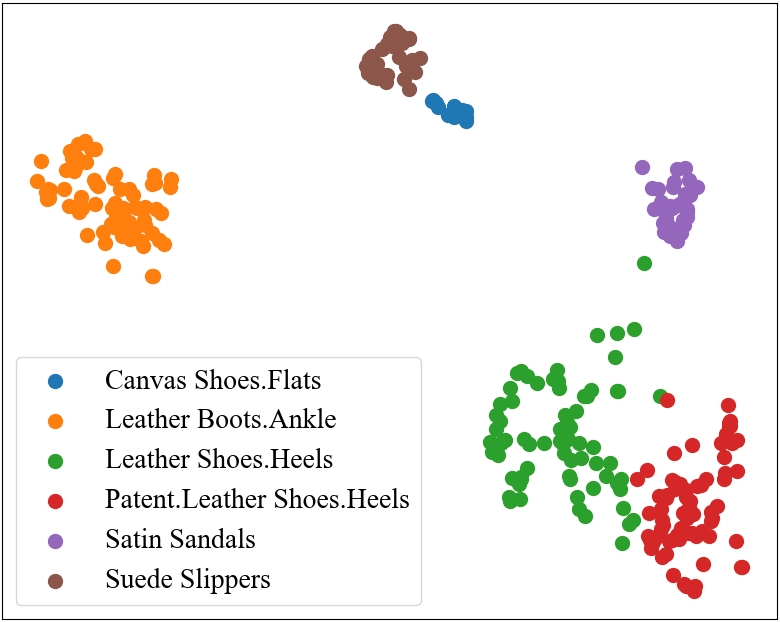}
    \caption{Compostion representations $f^c$.}
    \label{fig:subA}
  \end{subfigure}
  \begin{subfigure}[b]{0.25\textwidth}
    \includegraphics[width=\linewidth]{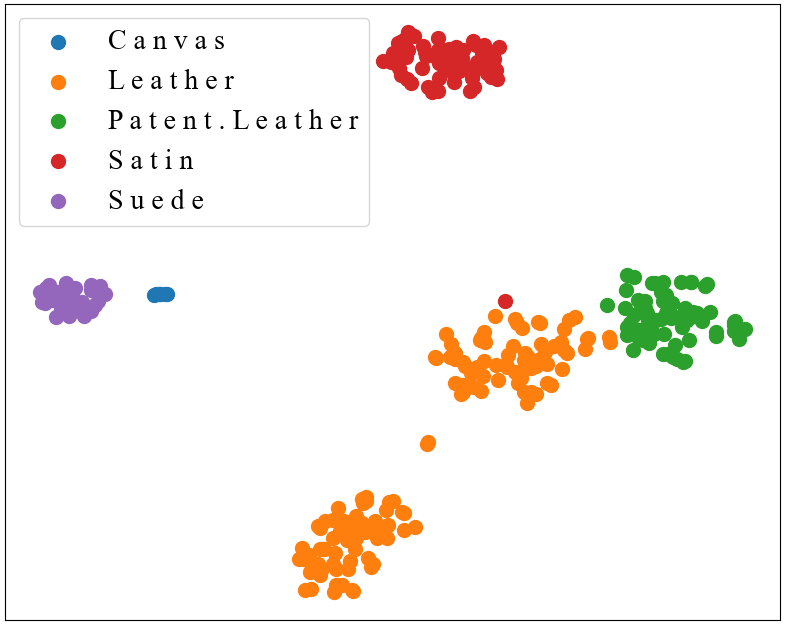}
    \caption{Attribute representations $f^a$.}
    \label{fig:subB}
  \end{subfigure}
  \begin{subfigure}[b]{0.25\textwidth}
    \includegraphics[width=\linewidth]{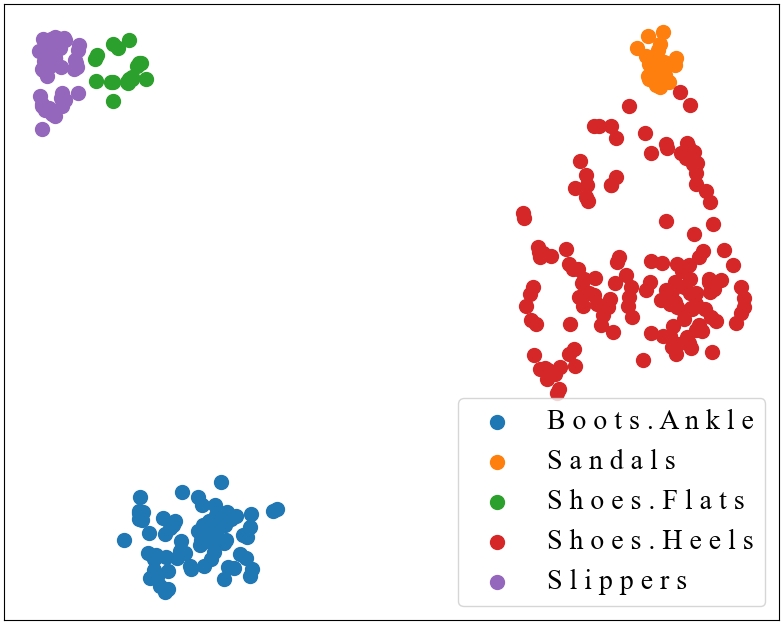}
    \caption{Object representations $f^o$.}
    \label{fig:subC}
  \end{subfigure}
    \begin{subfigure}[b]{0.25\textwidth}
    \includegraphics[width=\linewidth]{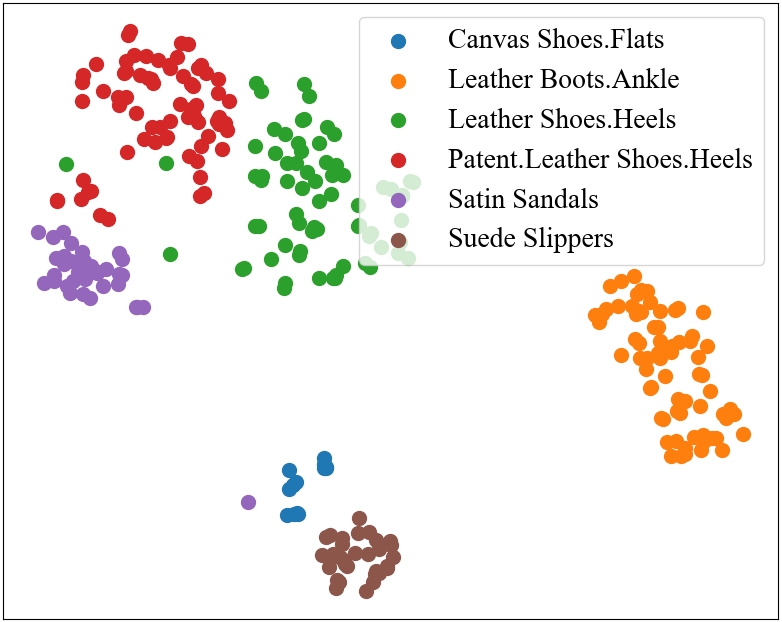}
    \caption{Baseline representations $f^g$.}
    \label{fig:subD}
  \end{subfigure}
    \begin{subfigure}[b]{0.25\textwidth}
    \includegraphics[width=\linewidth]{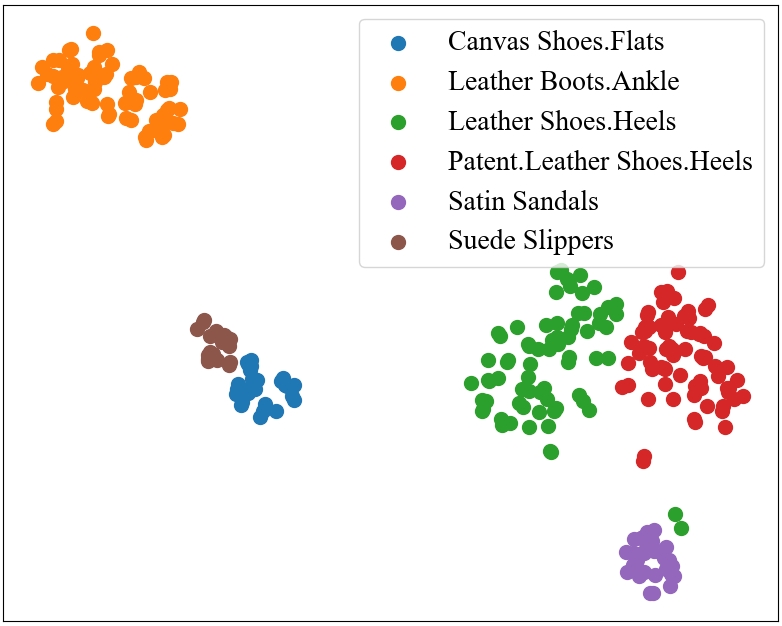}
    \caption{Troika representations.}
    \label{fig:subE}
  \end{subfigure}
  \begin{subfigure}[b]{0.25\textwidth}
    \includegraphics[width=\linewidth]{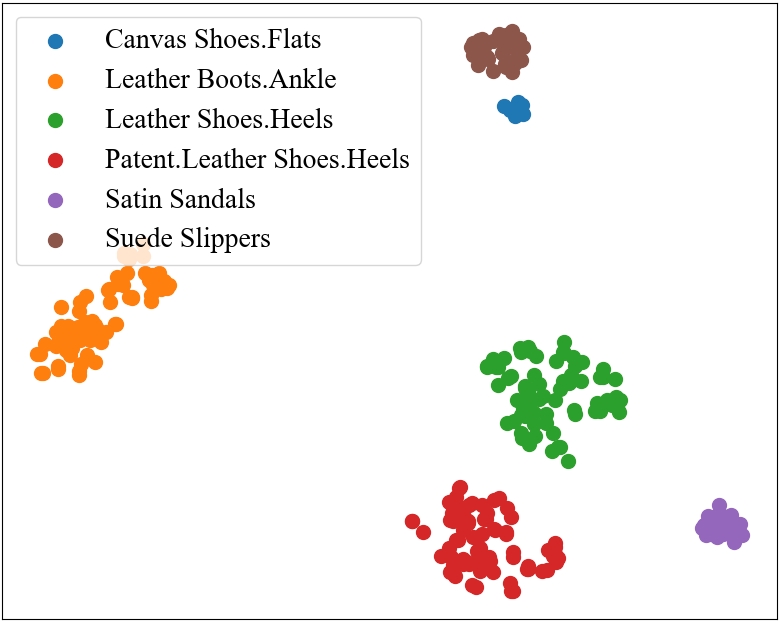}
    \caption{Combined representations (Ours).}
    \label{fig:subF}
  \end{subfigure}
  \caption{Visualization of baseline representations $f^g$, composition representations $f^c$, attribute representations $f^a$, object representations $f^o$, and combined representations learned by \model \ and Troika representations on UT-Zappos.}
  \label{fig:row}
      \label{fig:3}
\end{figure*}
\section{Experiments}
\subsection{Experimental Setup}
\paragraph{Datasets.}
We evaluate \model\ on three CZSL benchmark datasets. MIT-States \cite{c23} contains 53,753 images of everyday scenes, covering 115 attributes and 245 objects. UT-Zappos50K \cite{c24} comprises 50,025 images of footwear products, focusing on fine-grained recognition across 12 footwear categories and 16 material attributes. C-GQA \cite{c8} comprises 39,298 images, encompassing a massive coverage of real-world common concepts through diverse object and attribute categories.
\paragraph{Metrics.}
Following previous studies \cite{c20,c12}, we adopt best-Seen accuracy (S), best-Unseen accuracy (U), best-Harmonic Mean (HM), and Area Under the Curve (AUC) as evaluation metrics in both closed-world and open-world settings, with AUC being the most prioritized for its comprehensive assessment. Since the test set includes both seen and unseen compositions, the presence of seen compositions as error labels can bias the recognition of unseen ones. To mitigate this, we follow \cite{c49} and introduce a calibration bias term ranging from $-\infty$ to $+\infty$ to balance seen-unseen accuracy: a positive bias increases accuracy on unseen compositions, while a negative bias favors seen compositions.
\paragraph{Implementation Details.}
We adopt the pre-trained CLIP ViT-L/14 model~\cite{c10} as the backbone for both the image and text encoders in \model. The image encoder is fine-tuned via Low-Rank Adaptation (LoRA)~\cite{c52} applied to the last $M$ layers, with all components implemented in PyTorch. Training and evaluating are performed on a single NVIDIA RTX 6000 GPU (48 GB memory) for 15 epochs on each of the three public benchmarks (MIT-States, UT-Zappos, and C-GQA), using the Adam optimizer~\cite{c48}. Additional details are provided in the supplementary material.
\paragraph{Baseline Architecture.}
To facilitate subsequent testing, we fine-tune only the last $M$ layers of the image encoder using LoRA, following CSP \cite{c12}, to obtain the baseline model. In our method, this serves as the global branch.
\subsection{Main Results}
We separately test closed-world and open-world performance of \model \ and compare it with SOTA methods using the same backbone (ViT-L/14).

 \cref{tab:1} shows the results of \model \ and other SOTA methods in \textbf{Closed-World} setting on the MIT-States, UT-Zappos, and C-GQA. \model \ achieves the best or second-best results across all metrics on all datasets, with AUC improvements of \emph{+1.7\%}, \emph{+1.7\%}, and \emph{+13.7\%} over the second-best method, respectively. on the C-GQA, it further outperforms in S, U, and HM by \emph{+2.4\%}, \emph{+6.9\%}, and \emph{+9.3\%}, owing to its fine-grained decoupling of attribute and object concepts, which enhances generalization to unseen compositions.
 
 We present \textbf{Open-World} setting results of \model\ in \cref{tab:2}, performance of SOTA methods drops notably, yet \model\ maintains top or near-top performance across all datasets, achieving AUC of \emph{9.1\%}, \emph{40.3\%}, and \emph{4.8\%} on MIT-States, UT-Zappos, and C-GQA, respectively, with standout fine-grained results on UT-Zappos (S: \emph{71.2\%}, U: \emph{66.4\%}, H: \emph{54.5\%}) and superior generalization on C-GQA over methods like Logic-Troika (AUC: 3.4\%), underscoring its robust ability to disentangle and recognize complex semantic compositions.
\subsection{Ablation Study}
In the following section, we report experimental results of the ablation study for our framework under the closed-world setting.
\begin{table}[h]
    \centering
    \caption{Ablation on Multi-Space Disentanglement of \model \ on the MIT-States and UT-Zappos in closed-world setting.}
    \setlength{\tabcolsep}{0.5mm}
    \begin{tabular}{lcccccccc}
        \toprule
        \multirow{2}{*}{Method} & \multicolumn{4}{c}{MIT-States} & \multicolumn{4}{c}{UT-Zappos} \\
        \cmidrule(lr){2-5} \cmidrule(lr){6-9}
        & S & U & HM & AUC & S & U & HM & AUC \\
        \midrule
        $g$ & 45.5 & 53.4 & 37.3 & 20.6 & 67.6 & 70.2 & 52.9 & 40.0 \\
        $a + o$ & 46.8 & 40.9 & 32.3 & 15.6 & 71.7 & 72.4 & 50.1 & 39.7 \\
        $g + a + o$ & 50.0 & 53.3 & 39.9 & 22.7 & 71.2 & 75.3 & 57.0 & 45.2 \\
        $c + a + o$ & 42.2 & 52.6 & 35.3 & 18.4 & 70.5 & 74.2 & 57.6 & 44.8 \\
        \rowcolor{\graylevel}
        $g + c + a + o$ & \textbf{53.0} & \textbf{54.3} & \textbf{41.0} & \textbf{24.2} & \textbf{71.6} & \textbf{76.6} & \textbf{58.5} & \textbf{47.4} \\
        \bottomrule
    \end{tabular}
    \label{tab:3}
\end{table}
\begin{table}[h]
    \centering
        \caption{Ablation on Gated Mechanism (GM) on the MIT-States and UT-Zappos.}
    \setlength{\tabcolsep}{0.5mm}
    \begin{tabular}{l cccccccc}
        \toprule
        \multirow{2}{*}{Component} & \multicolumn{4}{c}{MIT-States} & \multicolumn{4}{c}{UT-Zappos} \\
        \cmidrule(lr){2-5} \cmidrule(lr){6-9} 
        & S & U & HM & AUC & S & U & HM & AUC \\
        \midrule
        w/o GM & 50.3 & 53.7 & 39.9 & 22.8 & 70.3  & 75.8 & 58.1 & 46.3 \\
        \rowcolor{\graylevel}
        w GM & \textbf{53.0} & \textbf{54.3} & \textbf{41.0} & \textbf{24.2} & \textbf{71.6} & \textbf{76.6} & \textbf{58.5} & \textbf{47.4}\\
        \bottomrule
    \end{tabular}
    \label{tab:4}
\end{table}
\begin{figure}[t]
\centering
\includegraphics[width=0.4\textwidth]{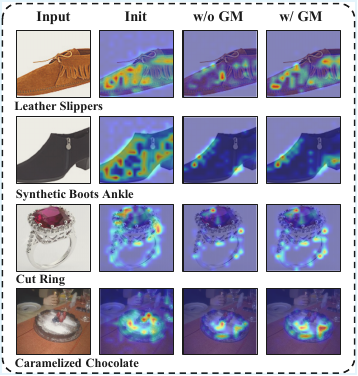} 
\caption{Visualization analysis of Gated Mechanism (GM). Init denotes the regions attended to by the latent units in the initial layer. The second and third columns respectively show the information attended to by the model in the final layer.}
\label{fig:4}
\end{figure}

\begin{figure}[t]
\centering
\includegraphics[width=0.45
\textwidth]{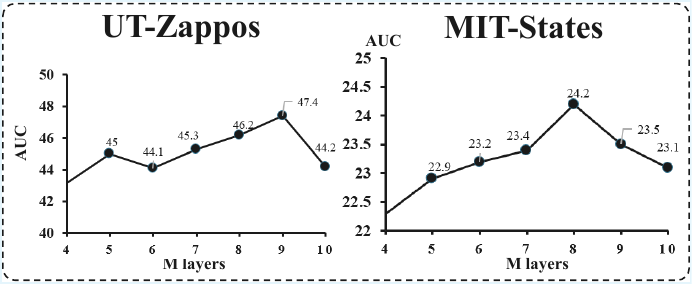}
\caption{Ablation on Layer Selection $M$.}
\label{fig:5}
\end{figure}

\paragraph{Ablation on Multi-Space Disentanglement.}
We conducted ablation experiments on MIT-States and UT-Zappos in a closed-world setting to validate each branch ($g$: global, $c$: composition, $a$: attribute, $o$: object). As shown in \cref{tab:3}, the results demonstrate that the branches of \model{} can work synergistically. The experimental results indicate that using baseline $g$, although capable of leveraging prior knowledge to achieve reasonable performance on MIT-States, has limited ability to handle fine-grained tasks and performs poorly on the UT-Zappos dataset. When $c+a+o$ are applied, the model’s performance on UT-Zappos improves significantly due to the fine-grained semantic extraction of visual features and the ability of the semantic feature decoupling module to disentangle attributes and objects. We further incorporated the $a+o$ into the baseline model $g$, resulting in a significant improvement in performance on both MIT-States and UT-Zappos. Ultimately, the complete model integrating all modules achieved the best performance. Compared with the baseline model, the proposed model achieved improvements of \emph{10.2\%} and \emph{17.4\%} in HM and AUC, respectively, on MIT-States, and remarkable improvements of \emph{12.0\%} (HM) and \emph{18.5\%} (AUC) on UT-Zappos. These results indicate that the modules designed in \model{} are not only effective individually but also capable of working synergistically to complement each other’s strengths.

To further validate the effectiveness of each branch, we visualized the distribution of visual representations using T-SNE \cite{c54}, as shown in \cref{fig:3}. In the global representation $f^g$ (\cref{fig:subD}), clusters like "Satin Sandals" heavily overlap and are hard to distinguish. In the composition representation $f^c$ (\cref{fig:subA}), "Satin Sandals" is clearly separated, with significantly increased inter-class distances. The attribute representation $f^a$ (\cref{fig:subB}) effectively distinguishes "Leather" from "Patent Leather", while the object representation $f^o$ enlarges inter-object gaps. When all branches are combined (\cref{fig:subF}), all compositions are clearly separable, outperforming Troika (\cref{fig:subE}). This demonstrates that each branch is rationally designed and complementary, significantly enhancing discriminative power.

\begin{table}[htbp]
\centering
\caption{Ablation on backbone architecture.}
\setlength{\tabcolsep}{1mm}
\footnotesize
\begin{tabular}{ll *{3}{c}}
\toprule
\multirow{2}{*}{\textbf{Method}} & \multirow{2}{*}{\textbf{Backbone}} &
\multicolumn{1}{c}{\textbf{MIT-States}} &
\multicolumn{1}{c}{\textbf{UT-Zappos}} &
\multicolumn{1}{c}{\textbf{C-GQA}} \\
\cmidrule(lr){3-3} \cmidrule(lr){4-4} \cmidrule(lr){5-5}
 & & AUC & AUC & AUC \\
\midrule
CLIP \cite{c10} & ViT-B/32 & 7.5 & 2.4 & 1.2 \\
CSP \cite{c12} & ViT-B/32 & 12.4 & 24.2 & 5.7 \\
Troika \cite{c11} & ViT-B/32 & 13.9 & 32.3 & 8.4 \\
CAILA \cite{c46}& ViT-B/32 & 16.1 & 39.0 & 10.4 \\
\rowcolor{\graylevel}
\textbf{\model\ (Ours)} & ViT-B/32 & \textbf{16.4} & \textbf{40.5} & \textbf{11.4} \\
\midrule
CLIP \cite{c10} & ViT-L/14 & 11.0 & 5.0 & 1.4 \\
CSP \cite{c12} & ViT-L/14 & 19.4 & 33.0 & 6.2 \\
Troika \cite{c11} & ViT-L/14 & 22.1 & 41.7 & 12.4 \\
CAILA \cite{c46}& ViT-L/14 & 23.4 & 44.1 & 14.8 \\
\rowcolor{\graylevel}
\textbf{\model\ (Ours)} & ViT-L/14 & \textbf{24.2} & \textbf{47.4} & \textbf{17.4} \\
\bottomrule
\end{tabular}
\label{tab:5}
\end{table}
\begin{table}[h]
    \centering
    \footnotesize
    \caption{Ablation on efficiency comparison on UT-Zappos.}
    \label{tab:6}
    \setlength{\tabcolsep}{1mm}
    \begin{tabular}{lcccc}
        \toprule
\textbf{Method} & \textbf{Memory}~$\downarrow$ & \textbf{Training}~$\downarrow$ & \textbf{Inference}~$\downarrow$ & \textbf{AUC}~$\uparrow$ \\
        \midrule
        Troika \cite{c11}&
          19.9G & 4.1min & 14.9ms & 41.7 \\
        CLUSPRO \cite{c17} &
          18.5G & 4.6min & 14.6ms & 46.6 \\
        \rowcolor{\graylevel}
          \textbf{\model \ (Ours)} &
          \textbf{14.4G} & \textbf{3.1min} & \textbf{13.7ms} & \textbf{47.4} \\
        \bottomrule
    \end{tabular}
\end{table}
\paragraph{Ablation on Gated Mechanism.}
To verify the effectiveness of the gating mechanism (GM) in the model, we conducted an ablation study on the MIT-States and UT-Zappos datasets. Specifically, we either retained or removed the GM and compared the model’s performance across different metrics, as shown in \cref{tab:4}. The results indicate that incorporating the GM improves the model’s performance on S, U, HM, and AUC metrics, with a particularly notable improvement in AUC. These findings suggest that, for this task, adding gating to the cross-attention mechanism enables the model to better learn visual features of attributes and objects, thereby enhancing overall performance. 

To qualitatively evaluate whether GM indeed facilitates the extraction of target-related semantic information, we visualized the attention weights of several test samples from the UT-Zappos dataset, as shown in \cref{fig:4}. It can be observed that with the introduction of GM, attention focuses more on regions semantically related to the target, highlighting GM’s role in feature extraction. Moreover, we can see that in the initial layer, the model mainly attends to low-level semantic cues of the target, while in the final layer, it tends to focus on more abstract semantic representations, thereby learning better target representations.
\paragraph{Ablation on backbone architecture.}
\cref{tab:5} compares the performance of \model \ with other methods when using different ViT-based CLIP backbones. It can be seen that \model \ performs well on both large and small backbones, demonstrating the effectiveness and generalizability of our method across different model scales.
\paragraph{Ablation on efficiency comparison.}
\cref{tab:6} shows efficiency comparison results of \model \ with state-of-the-art methods. Compared to Troika and CLUSPRO, our method demonstrates significant advantages in training time, GPU memory usage, inference speed, and classification performance, fully highlighting the comprehensive superiority of our method in balancing efficiency and effectiveness.
\subsection{Qualitative analysis}
\begin{figure}[t]
\centering
\includegraphics[width=0.45\textwidth]{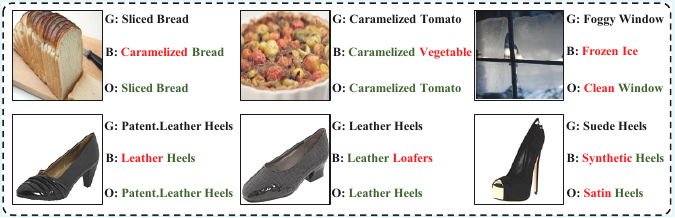} 
\caption{Qualitative analysis on UT-Zappos and MIT-States. Here, G denotes the ground truth, B denotes the baseline model, and O denotes our method \model, with correct predictions marked in green and incorrect predictions marked in red.}
\label{fig:6}
\end{figure}

\cref{fig:6} illustrates \model’s predictions on several images. Benefiting from fine-grained feature decoupling, the model can accurately distinguish subtle differences such as “Patent Leather” versus “Leather” in UT-Zappos, as well as highly entangled combinations like “Caramelized Tomato” in MIT-States. In the failure cases shown in the last column, \model \ is still able to partially recognize the relevant features, demonstrating a certain level of robustness.
\section{Conclusion}
In this paper, we propose \model, which uses Gated Cross-Attention to extract fine-grained semantic features. To disentangle attribute and object concepts, we introduce Multi-Space Disentanglement, which represents attributes and objects in multidimensional spaces. Experiments on three benchmark datasets show that our method significantly outperforms existing methods across all metrics. In the future, we will refine fine-grained disentanglement techniques and incorporate multimodal knowledge graphs to enhance semantic associations, thereby improving the model's generalization ability in complex compositional scenarios.

{
    \small
    \bibliographystyle{ieeenat_fullname}
    \bibliography{main}
}

\clearpage
\setcounter{page}{1}
\maketitlesupplementary

\appendix
\renewcommand{\thesection}{\Alph{section}}

\section{Implementation Details}
In \model, we use the pre-trained and frozen CLIP ViT-L/14  model as the backbone for both the image and text encoders. The last M layers of the image encoder are fine-tuned via LoRA. Training and evaluation are conducted on a single NVIDIA RTX 6000 GPU (48 GB memory), and all components are implemented in PyTorch. \cref{tab:7} lists the hyperparameters used for the three datasets.

\begin{table}[htbp]
\centering
\caption{Hyperparameters for MIT-States,UT-Zappos,C-GQA.}
\setlength{\tabcolsep}{1mm}
\footnotesize
\begin{tabular}{l *{3}{c}}
\toprule
\textbf{Hyper parameters} &
\multicolumn{1}{c}{\textbf{MIT-States}} &
\multicolumn{1}{c}{\textbf{UT-Zappos}} &
\multicolumn{1}{c}{\textbf{C-GQA}} \\
\midrule
BatchSize & 64 & 64 & 64 \\
Learning Rate & $10^{-4}$ & $2.5\times10^{-4}$ &  $10^{-4}$ \\
Epochs & 15 & 15 & 15 \\
Scheduler & StepLR & StepLR & StepLR \\
Weight Decay & $10^{-5}$ & $10^{-5}$ & $10^{-5}$ \\
Latent Units & 32  & 32 & 32 \\
$M$  & 8 & 9 & 12 \\
Optimizer & Adam & Adam & Adam \\
$\beta$ &  0.85 & 0.65 & 0.85 \\
LoRA Bottleneck Dimension & 128 & 64 & 128 \\
LoRA Dropout & 0.1 & 0.1 & 0.1 \\
Coefficients $\alpha_a$,$\alpha_o$,$\alpha_c$,$\alpha_g$ &  0.5,0.5,1,1&  0.5,0.5,1,1 &  0.5,0.5,1,1\\
Layers of $\operatorname{Tr_a} ,\operatorname{Tr_o},\operatorname{Tr_c}$ &  1,1,1&   1,1,1& 1,1,1\\
Attribute Dropout & 0.3 & 0.3 & 0.3 \\
\bottomrule
\end{tabular}
\label{tab:7}
\end{table}

\section{Dataset Details}
We evaluated FGSD-CZSL on three CZSL benchmark datasets: MIT-States, UT-Zappos and C-GQA.

C-GQA \cite{c8}. The dataset comprises 39,298 images, encompassing a massive coverage of real-world common concepts through diverse object and attribute categories. It provides annotations for 7,767 distinct attribute-object compositions, establishing it as the largest benchmark dataset currently available within the field of CZSL.

UT-Zappos \cite{c24}. The dataset comprises 50,025 images of footwear products, focusing on fine-grained recognition across 12 footwear categories and 16 material attributes. A key challenge lies in the fact that materials such as “synthetic leather” and “genuine leather” exhibit highly similar visual features, requiring the model to capture subtle texture and material differences.

MIT-States  \cite{c23}. The dataset contains 53,753 images of everyday scenes, covering 115 attributes and 245 objects, forming a total of 1,962 attribute-object composition. The data was primarily collected automatically using early image search engines, supplemented by limited manual annotation. Due to significant label noise, this dataset imposes higher demands on model robustness.

\section{Comparison Methods}
To evaluate our FGSD-CZSL method, we conducted comparative experiments with several state-of-the-art CZSL methods, including:

CSP \cite{c12} uses CLIP in the CZSL task, replacing category names in text prompts with trainable attribute and object labels for compositional recognition;
 
DFSP \cite{c15} constructs attribute-object joint representations via soft prompts and performs cross-modal decomposition and fusion in the language feature space, effectively enhancing recognition of unseen compositions;
 
CAILA \cite{c46} inserts attribute, object, and composition adapters into each CLIP layer and fuses features via MoA, enabling intra-layer concept-aware fine-tuning;
 
CDS-CZSL \cite{c18} quantifies attribute informativeness by evaluating applicability and contextual relevance across objects, selecting more effective attribute-object compositions for training;

Troika \cite{c11} designs a cross-modal traction module to align path-specific prompt representations with visual features, mitigating multi-modal representation biases and improving composition recognition;

PLID \cite{C30} leverages large language models to generate diverse, information-rich category distributions and dynamically fuses classification results from the composition and primitive spaces via a visual-language primitive decomposition module;

RAPR \cite{c14} retrieves relevant attribute and object representations from a training image database to identify unseen compositions, providing a retrieval-enhanced recognition strategy;

MSCI \cite{c19} employs a two-stage interaction mechanism to extract low-level local and high-level global features from the CLIP visual encoder, dynamically adjusting attention in text representations to enhance fine-grained attribute-object perception;

LOGICZSL \cite{c51} formalizes attribute-object relationships mined from large language models into logical rules and injects them via a logic-guided loss, explicitly modeling rich compositional semantics;

CLUSPRO \cite{c17} performs online clustering for each attribute/object to discover multiple diversified prototypes, leveraging them with contrastive and decorrelation constraints to improve compositional generalization.
\end{document}